\documentclass{egpubl}
\usepackage{eurova2025}
 
\WsPaper           

\CGFccby

\usepackage[T1]{fontenc}
\usepackage{dfadobe}  

\biberVersion
\BibtexOrBiblatex
\usepackage[backend=biber,bibstyle=EG,citestyle=alphabetic,backref=true]{biblatex} 
\electronicVersion
\PrintedOrElectronic

\ifpdf \usepackage[pdftex]{graphicx} \pdfcompresslevel=9
\else \usepackage[dvips]{graphicx} \fi

\usepackage{egweblnk} 

\usepackage{eso-pic}

\AddToShipoutPicture{%
  \AtPageUpperLeft{%
    \raisebox{-1.5cm}{\makebox[210mm][c]{\custompreprinttext}}%
  }%
}

\usepackage{amsfonts}
\usepackage{amsmath}
\usepackage{booktabs}
\usepackage{capt-of}
\usepackage{enumitem}
\usepackage{float}
\usepackage{microtype}
\usepackage{wrapfig}

\newcommand{\papertitle}{Evaluating Autoencoders for Parametric and \\ Invertible Multidimensional Projections}

\AtBeginDocument{

}

\newcommand{\subhead}[1]{
  \noindent\textbf{#1}%
}

\newcommand{\compresslist}{
  \setlength{\itemsep}{1pt}
  \setlength{\parskip}{0pt}
  \setlength{\parsep}{0pt}	
}

\DeclareMathAlphabet{\mathpzc}{OT1}{pzc}{m}{it}

\newcommand{\verticalLabel}[1]{\rotatebox{90}{\parbox{0.185\textwidth}{\centering\textbf{#1}}}}

\newcommand{\verticalLabelSmall}[1]{\rotatebox{90}{\parbox{1.1cm}{\centering\textbf{#1}}}}

\newcommand{\horizontalLabel}[1]{\parbox{0.21\textwidth}{\centering\textbf{#1}}}

\newcommand{\custompreprinttext}{%
\parbox{156mm}{%
\itshape
$\copyright$ 2025~The Authors. This is the author's version of the article that was published in the proceedings of the 16th International EuroVis Workshop on Visual Analytics, available at: \href{https://doi.org/10.2312/eurova.20251099}{\color{blue}10.2312/eurova.20251099}%
}}

\title[Evaluating Autoencoders for Parametric and Invertible Multidimensional Projections]%
      {\papertitle}

\author[Dennig et al.]
{\parbox{\textwidth}{\centering
        Frederik L. Dennig$^1$\orcid{0000-0003-1116-8450},
        Nina Geyer$^1$\orcid{0009-0000-9876-0582},
        Daniela Blumberg$^1$\orcid{0009-0002-2090-9847},
        Yannick Metz$^1$\orcid{0000-0001-5955-4487},
        and Daniel A. Keim$^1$\orcid{0000-0001-7966-9740}
        }
        \\
{\parbox{\textwidth}{\centering $^1$University of Konstanz, Germany
       }
}
}

\begin{document}


\maketitle
\begin{abstract}
   Recently, neural networks have gained attention for creating parametric and invertible multidimensional data projections. Parametric projections allow for embedding previously unseen data without recomputing the projection as a whole, while invertible projections enable the generation of new data points. However, these properties have never been explored simultaneously for arbitrary projection methods. We evaluate three autoencoder (AE) architectures for creating parametric and invertible projections. Based on a given projection, we train AEs to learn a mapping into 2D space and an inverse mapping into the original space. We perform a quantitative and qualitative comparison on four datasets of varying dimensionality and pattern complexity using t-SNE. Our results indicate that AEs with a customized loss function can create smoother parametric and inverse projections than feed-forward neural networks while giving users control over the strength of the smoothing effect.


\end{abstract}  

\section{Introduction}

Multidimensional projections, also known as dimensionality reduction (DR) methods, are well-established and frequently used to analyze high-dimensional data visually \cite{Jolliffe1986}.
DR methods reduce high-dimensional data to a lower-dimensional projection, usually to 2D or 3D, while trying to preserve relationships, i.e., distances and neighborhoods \cite{Nonato2019}.
A shortcoming of these approaches is that the result can be hard to interpret since the relationships are projected non-linearly.  
To address this, researchers proposed to enhance the visualization through layout enrichments, showing the distortions and gradients between projected points \cite{Nonato2019, Espadoto2021Unprojection, Dennig2024}. 
However, we argue that these methods benefit from projections that are \emph{parametric} and \emph{invertible}.
(1) Parametric methods use a model with learnable parameters to control the mapping between high- and low-dimensional spaces. Doing so allows the projection of new data points, real and synthetic, without recalculating all pairwise relationships. This is faster and keeps the projection stable \cite{Sainburg2021}.
(2) Invertible methods have a smooth mapping from the projection space back to the original high-dimensional space, allowing us to generate new data from any arbitrary position of the projection \cite{Wijk2003}, e.g., for interactive counterfactual generation \cite{Schlegel2024}.
However, DR methods, such as \textit{t-SNE} \cite{Maaten2009}, are generally not parametric or invertible.
In recent years, neural networks (NNs) have been used to learn mappings from high-dimensional space to the projection space \cite{Espadoto2020Deep}, and vice versa \cite{Espadoto2021Unprojection, Hinterreiter2023}.
However, while these approaches can be applied to arbitrary projections, they only consider each mapping direction individually.
In this work, we explore and evaluate different autoencoders (AEs) \cite{Bank2023} to support parametric mapping and inversion jointly through their encoder-decoder architecture.
More specifically, we compare the individual \textit{feed-forward NNs} proposed by Espadoto et al.~\cite{Espadoto2021Unprojection} and Appleby et al.~\cite{Appleby2022} with \textit{standard} AEs and \textit{variational} AEs with custom loss functions for representing the projection space in the latent space. 
We measure their ability to parametrically project and inverse project data points and compare them qualitatively by visually assessing their outputs.
Additionally, we analyze the smoothness of the projection through gradient maps.
Overall, we contribute the following:
\begin{enumerate}[label=(\arabic*),left=0pt]
\compresslist
\item We propose three \emph{AE-based NN architectures} for creating parametric and invertible projections.
\item We perform an \textit{evaluation} comparing the three architectures \textit{quantitatively} and \textit{qualitatively} on four datasets using t-SNE.
\item For \textit{reproducibility}, we provide the analysis, results and source code at \url{https://osf.io/r2yqd}.
\end{enumerate}

With this work, we aim to enhance high-dimensional data analysis by improving the interpretation and explainability of DR methods.

\section{Related Work on Multidimensional Projections}

Let $D = \{x_{i}\}_{1 \leq i \leq n}$ be a high-dimensional dataset with $d$ dimensions and $n$ samples $x_i \in \mathbb{R}^d$.
A \emph{projection} method $P$ maps $D$ to $P(D) = \{P(x_i) | x_i \in D\} = \{y_i\}_{1 \leq i \leq n}$, where $P(D) \subset \mathbb{R}^q$ with  $q \ll d$.
For our case, $q = 2$; thus, $P(D)$ can be visualized in a 2-dimensional scatterplot.
Projection methods have been extensively studied and assessed in several surveys \cite{Yin2007, Cunningham2015, Espadoto2019, Nonato2019}.
They can be classified into \emph{linear} and \emph{non-linear} methods~\cite{Shusen2017, Espadoto2019}.
Additionally, they either preserve \emph{global} structure or \emph{local} neighborhoods.
Linear projection methods like PCA \cite{Jolliffe1986} can be computed very efficiently and preserve the global structure of the data. MDS \cite{kruskal1978multidimensional} is an example of a global, non-linear projection method.
There exist many \emph{non-linear} methods that put stronger weight on preserving \emph{local} neighborhood structures at the expense of global structure fidelity.
Such methods include t-SNE \cite{Maaten2008Tsne} or UMAP \cite{McInnes2018}.
Autoencoder-based approaches generally also fall into this category \cite{wang2016auto}.

\subhead{Parametric Projections Methods:}
Standard non-linear DR methods \cite{kruskal1978multidimensional, Maaten2008Tsne, McInnes2018} are non-parametric, needing complete recalculation when projecting new data points \cite{Hinterreiter2023}.
Parametric approaches \cite{Bunte2012} address this limitation, e.g., by using NNs to learn mapping functions into lower-dimensional space \cite{Hinton2006, Maaten2009}.
Van der Maaten~\cite{Maaten2009} used a feed-forward NN to build \emph{parametric t-SNE}.
Similarly, \emph{parametric UMAP} \cite{Sainburg2021} uses NNs, including autoencoders, designed to replace the non-parametric embedding step.
By training NNs to infer 2D coordinates for input domain data points, Espadoto et al.~\cite{Espadoto2020Deep} showed that NNs with sufficient size can effectively approximate many existing non-parametric methods.
\emph{HyperNP} \cite{Appleby2022} uses NNs to approximate projection techniques across hyperparameters (e.g., perplexity for t-SNE), allowing for their interactive exploration.
\textit{ParaDime} \cite{Hinterreiter2023} streamlines the creation of NN-based DR approaches by proposing a grammar specific to parametric DR.

\subhead{Inverse Projections:}
\emph{Inverse projections} are functions $P^{-1}: \mathbb{R}^q \rightarrow \mathbb{R}^d$ which are smooth and minimize the cost $\sum_{x \in D} \|P^{-1}(P(x)) - x\|$ with projection $P$ and $\| \cdot \|$ denoting the $L_2$ norm.
Only a handful of projections are inherently inversely defined, e.g., UMAP \cite{McInnes2018}.
An inverse mapping is not explicitly available for most other projection methods, requiring alternative methods, such as NNs  \cite{Espadoto2021Unprojection}.
One of the early methods introduced a global interpolation-based technique to generate new data points \cite{Wijk2003}.
Later, \emph{iLAMP} \cite{Amorim2012}, which builds on LAMP \cite{Joia2011}, addresses inverse projection by focusing on preserving local relationships. 
Amorim et al.~\cite{Amorim2015} later refined this by using a radial basis function kernel for interpolation.
Blumberg et al.~\cite{Blumberg2024} invert MDS projections by leveraging geometrical relationships between data points through multilateration.
Recently, deep learning was employed for creating inverse projections.
The feed-forward NNs proposed by Espadoto et al.~\cite{Espadoto2021Unprojection} learn the mapping from the projection space back to the high-dimensional space.

\section{Training Autoencoders for Multidimensional Projections}

We train AEs to \textit{jointly} create a parametric projection $P$ and inverse projection $P^{-1}$.
AEs are a class of NNs designed for unsupervised learning of efficient data representations \cite{Bank2023}.
AEs were already well-established in the context of DR \cite{Hinton2006}.
However, a standard AE will learn a latent space representation optimized for data compression and feature extraction.
Thus, modified loss functions were proposed to impose a structure on the latent space of an AE, e.g., SSNP~\cite{Espadoto2021Ssnp} and ShaRP~\cite{Machado2024} integrate pseudo-labels from clustering into their loss functions.
However, these approaches do not allow for inverting arbitrary user-defined projections.

\noindent
\begin{minipage}[c]{0.68\linewidth}
\includegraphics[width=\linewidth]{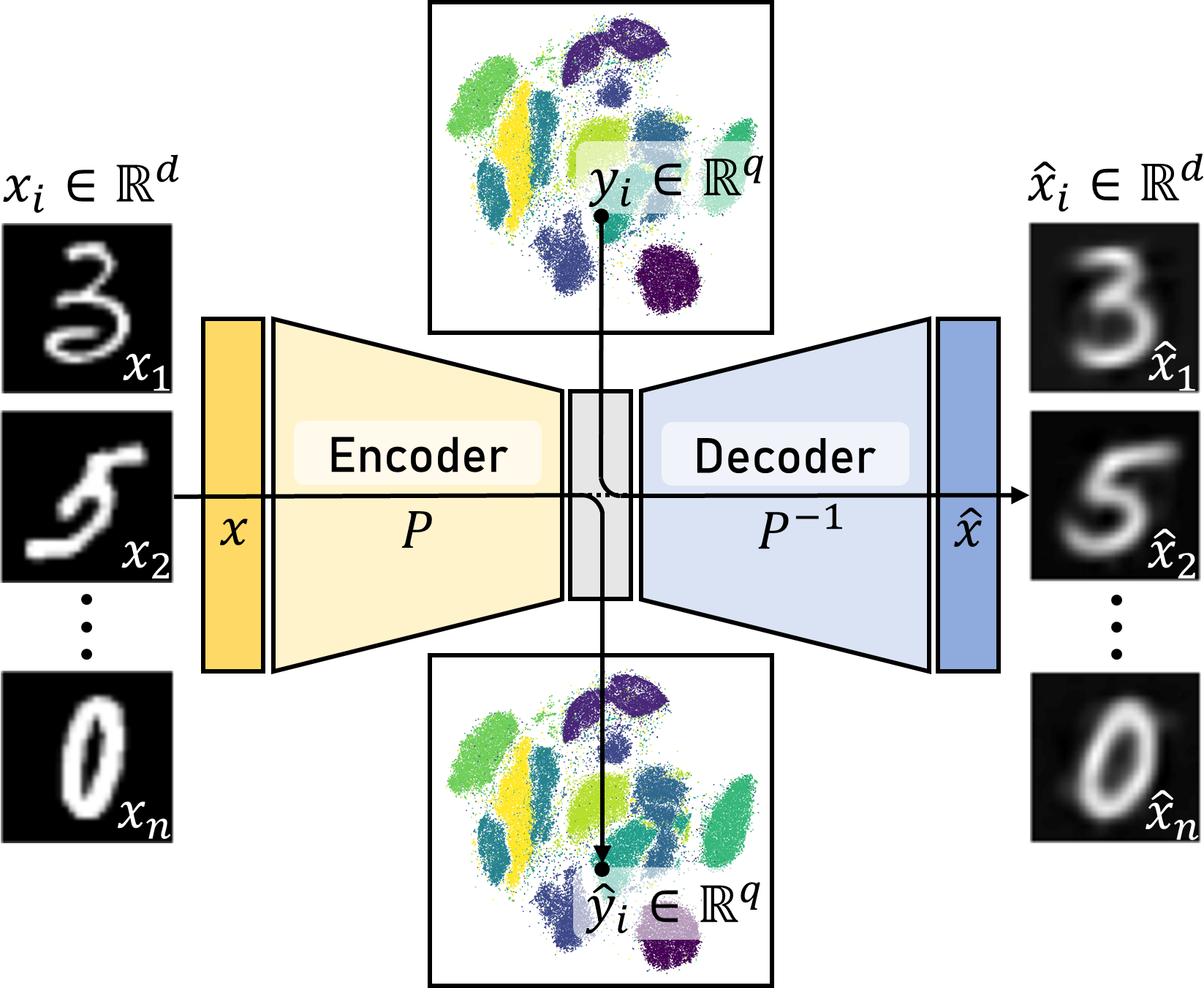}
\end{minipage}
\hfill
\begin{minipage}[c]{0.28\linewidth}
\captionof{figure}{The encoder of the AE learns the parametric projection $P$, mapping new data record $x_i$ into 2D space (as $\hat{y}_i$). The decoder learns the inverse projection $P^{-1}$ generating a high-dimensional sample $\hat{x}_i$ from any 2D point $y_i$.}
\label{fig:ae-overview}
\end{minipage}

An AE consists of an \emph{encoder} and a \emph{decoder} (\autoref{fig:ae-overview}).
The encoder $\mathrm{Enc}: \mathbb{R}^d \rightarrow \mathbb{R}^q$ learns a mapping from the high-dimensional input space into a lower-dimensional latent space $\hat{Y} = \{\hat{y}_i\}_{i=1}^n$, with $\hat{y}_i = \mathrm{Enc}(x_i) \in \mathbb{R}^q$ and $n$ data points.
In contrast, the decoder $\mathrm{Dec}: \mathbb{R}^q \rightarrow \mathbb{R}^d$ aims to map these latent representations back into the original input space $\hat{x}_i = \mathrm{Dec}(y_i) \in \mathbb{R}^d$.
A standard AE aims to learn a compressed latent encoding $y_i$ while reconstructing the input $x_i$ as accurately as possible.
During training, we feed the input $x_i$ forward through the encoder and decoder to obtain a reconstruction $\hat{x}_i$, then compute a reconstruction loss, i.e., \textit{end-to-end}.
A common choice for this loss is the mean squared error (MSE)
$\mathcal{L}_{rec}(x_i,\hat{x}_i) = MSE(x_i,\hat{x}_i) := \|x_i - \hat{x}_i\|^2$.
This error is \emph{backpropagated} through the AE, and its parameters are updated accordingly. 
We use modern training methods for \emph{deep} AEs, including batches of data and stochastic gradient descent~\cite{Kingma2015}.
We use standard loss functions, like the mean squared error (MSE) and the Kullback-Leibler divergence ($D_\textrm{KL}$), a measure of the difference of two probability distributions $A$ and $B$.
In the following paragraph, we describe our proposed NN architectures.

\begin{wrapfigure}[11]{r}{2.2cm} 
    \vspace*{-3.5mm} 
    \hspace*{-4.5mm}
    \includegraphics[width=2.5cm]{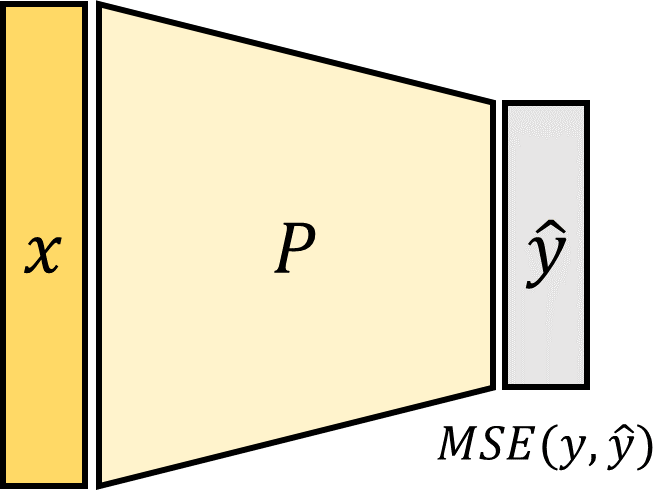} \\
    \hspace*{-4.5mm}
    \includegraphics[width=2.5cm]{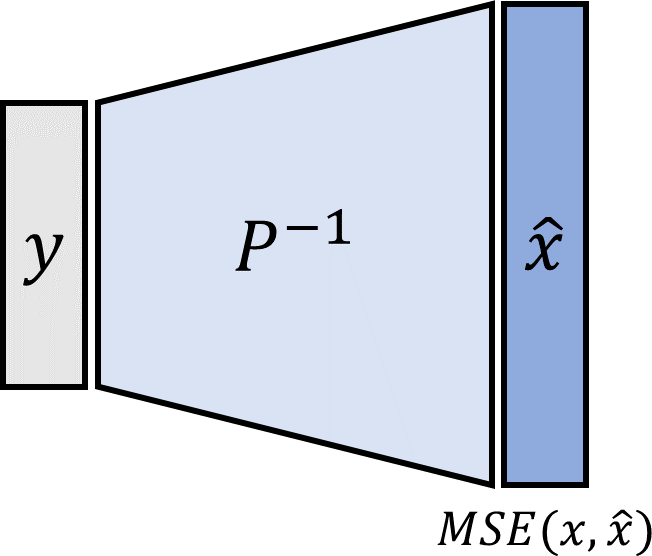}
\end{wrapfigure}
\subhead{Individual Projector and Reconstructor (P\&R):}
We train two standard feed-forward NNs to combine the approaches by Appleby et al. \cite{Appleby2022} (denoted as $P$) and Espadoto et al. \cite{Espadoto2021Unprojection} (denoted as $P^{-1}$).
This structure is, strictly speaking, not an AE but provides a baseline for our experiments.
We train a \textit{projector} network learning the parametric projection $P$, enabling the addition of new data to the projection through its loss function $\mathcal{L}_{pro}(x, \hat{y}) = MSE(P(x), \hat{y})$.
Independently, we train a \textit{reconstructor} network learning to invert the projection space via the loss function  $\mathcal{L}_{rec}(y, \hat{x}) = MSE(P^{-1}(y), \hat{x})$.
During training, $P$ and $P^{-1}$ are learned by the individual networks, \textit{projecting} and \textit{reconstructing} the dataset, i.e., $P(x) = y$ and $P^{-1}(y) = x$.

\begin{wrapfigure}[6]{r}{4.4cm} 
    \vspace*{-4.5mm} 
    \hspace*{-4.5mm}
    \includegraphics[width=4.7cm]{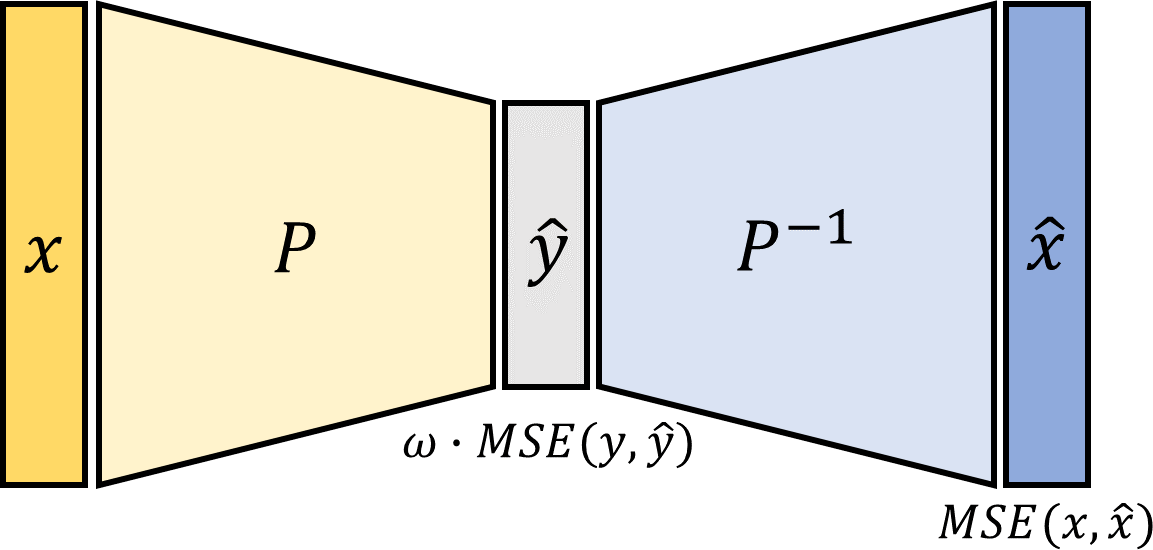}
\end{wrapfigure}
\subhead{Autoencoder with Latent Loss Term (AEL):}
To avoid potential projection and reconstruction artifacts caused by training individual networks, we propose using AEs.
In AEs, the encoder and decoder are trained jointly as one NN.
The encoder of the AE learns to project the data points, while the decoder learns the inverse projection.
To impose a structure on the latent space, we define the loss:
\begin{equation}\label{eq:loss-ael}
\mathcal{L}_{AEL}(x,\hat{x}, y, \hat{y}) \;=\; \textrm{MSE}(x,\hat{x}) + \omega \cdot \textrm{MSE}(y,\hat{y})
\end{equation}
Here, the $\textrm{MSE}(x,\hat{x})$ denotes the reconstruction loss.
To enable the AE to learn a given projection, we modify its loss function by adding a component $\omega \cdot \textrm{MSE}(y,\hat{y})$  to force its latent space to conform.
The weight $\omega \in \mathbb{R}^+$ determines the strength of the latent space to conform to the projection, i.e., between the projection and reconstruction quality. 
We performed a parameter scan (see supplementary material) and determined that $\omega = 0.5$ achieves a generally low MSE for test data.
We discuss the effect of $\omega$ in \autoref{sec:discussion}. 

\begin{wrapfigure}[6]{r}{4.8cm} 
    \vspace*{-4mm} 
    \hspace*{-4.5mm}
    \includegraphics[width=5.05cm]{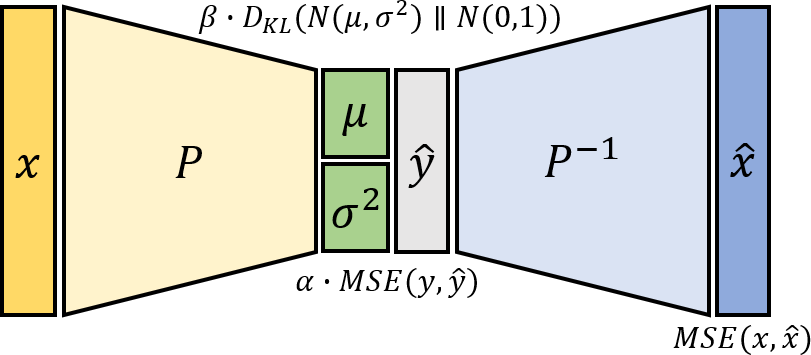}
\end{wrapfigure}
\begin{figure}[t]
\begin{minipage}[t]{0.23\linewidth}
\centering
\includegraphics[width=\linewidth]{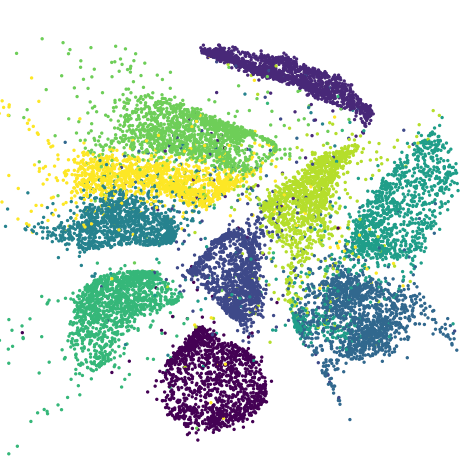} \\
(a) $\beta = 0.01$
\end{minipage}
\hfill
\begin{minipage}[t]{0.23\linewidth}
\centering
\includegraphics[width=\linewidth]{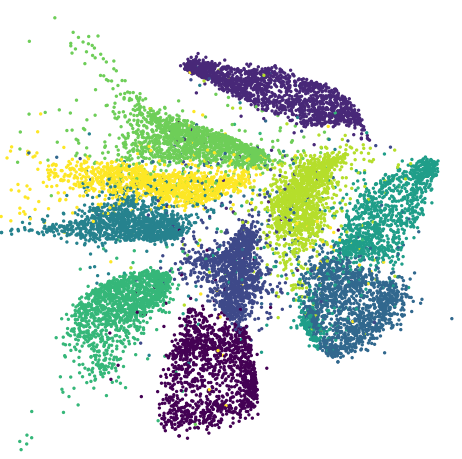} \\
(b) $\beta = 0.1$
\end{minipage}
\hfill
\begin{minipage}[t]{0.23\linewidth}
\centering
\includegraphics[width=\linewidth]{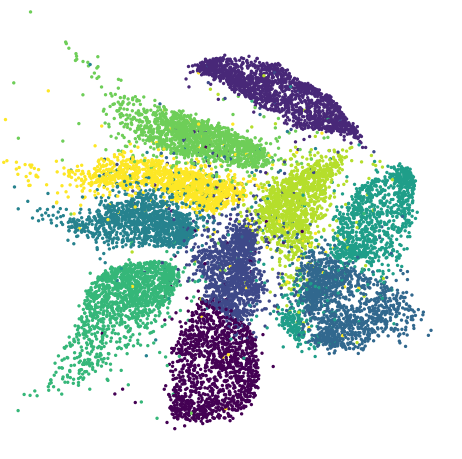} \\
(c) $\beta = 0.2$
\end{minipage}
\hfill
\begin{minipage}[t]{0.23\linewidth}
\centering
\includegraphics[width=\linewidth]{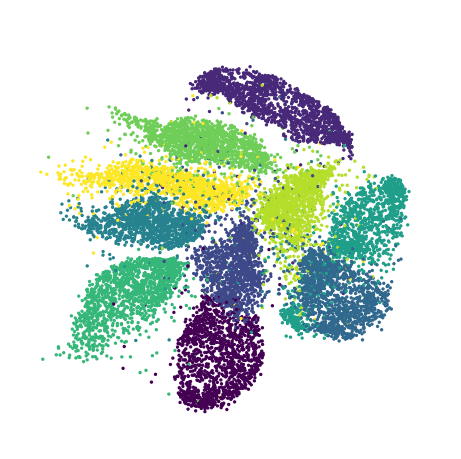} \\
(d) $\beta = 0.3$
\end{minipage}
\caption{The effect of $\beta$ on the parametric projection of VAEL using a t-SNE projection of MNIST. $\alpha$ is set to $1.0$ in all examples.}
\label{fig:beta-parameter}
\vspace*{-2em}
\end{figure}

\subhead{Variational Autoencoder with Latent Loss Term (VAEL):}
Instead of directly predicting the latent variable $\hat{y}$, the encoder predicts the parameters of the normal distribution $\mu$ and $\sigma^2$.
In training, $\hat{y}$ is sampled from a 2D normal distribution, i.e., $\hat{y} \sim \mathcal{N}(\mu, \sigma^2)$ with $\hat{y}$, $\mu$, $\sigma^2 \in \mathbb{R}^q$.
The VAE is still trained end-to-end, using the \emph{reparameterization trick}, to backpropagate through the sampling operation~\cite{Kingma2014}. For our VAEL, we incorporate the \emph{evidence lower bound loss (ELBO)} used for VAE training as:
\begin{align}
\begin{split}
\mathcal{L}_{VAEL}(x, \hat{x}, y, \mu, \sigma^2) = \textrm{MSE}&(x,\hat{x}) + \alpha \cdot \textrm{MSE}(y, \hat{y} \sim \mathcal{N}(\mu,\sigma^2)) \\
&+ \beta \cdot D_\textrm{KL}(\mathcal{N}(\mu,\sigma^2)\ ||\ \mathcal{N}(0,1))
\end{split}
\end{align}
We follow the framework of the $\beta-VAE$~\cite{Higgins2017}, which adds an additional $\beta$-parameter to balance conformity to the prior normal distribution and reconstruction quality. We sampled various combinations and found that $\alpha = 1.0$ and $\beta = 0.1$ achieve generally good quality in terms of MSE on test data. The effect of varying $\beta$ is shown in \autoref{fig:beta-parameter}.  We discuss the selection of $\alpha$ and $\beta$ in \autoref{sec:discussion}. Similar to Chen et al.~\cite{Chen2024}, we use $\mu$ as a 2D coordinate to create parametric projections, i.e., for inference only, since it is the mean and mode of the normal distribution. 

\subsection{Data Preprocessing and Training}

We derive the topologies of our AEs from the configurations of the NNs proposed by Espadoto et al., and Appleby et al.~\cite{Espadoto2021Unprojection, Appleby2022}.
For P\&R, we use these topologies directly; for AEL and VAEL, we combine the NNs to create a bottleneck according to our descriptions.
Using the \emph{Adam optimizer} \cite{Kingma2015}, we observed that the training of our AEs converged similarly fast.
Thus, we set the number of training epochs to 50, allowing all to converge.
The learning rate is set to 0.001, and the batch size is set to 32, which are the typical values.
We use batch normalization after each layer and dropout regularization with a probability of $0.25$.
For all details, refer to the source code.
In our experiments, we standardize the input data per dimension, i.e., the dataset and the projection.
When projecting or reconstructing data using the encoder or decoder, we apply the inverse of the standardization to the output, enabling us to recover the representation in the original space and the projection space, respectively.
This is possible since standardization is an invertible linear transformation due to $\sigma^2 \in \mathbb{R}^+$.

\section{Evaluation}

We evaluate the approaches quantitatively using the \textit{mean squared error (MSE)}, the \textit{average gradient} of the inverse projection, and \textit{runtime measurements}.
Additionally, we qualitatively compare the results of the NN architectures by visually comparing parametric and inverse projections.
Since there is no ground truth for many points $\hat{y} \in \mathbb{R}^2 \setminus P(D)$, we use \textit{gradient maps} \cite{Espadoto2021Unprojection} to evaluate the inverse projection for those points. 
For each $\hat{y}$, we compute a pseudo-derivative of $P^{-1}$ measuring the difference between the horizontal and vertical neighbors of a pixel defined as
$G(\hat{y}) = \sqrt { \|P^{-1}(\hat{y}_{left}) - P^{-1}(\hat{y}_{right})\|^2 + \|P^{-1}(\hat{y}_{up}) - P^{-1}(\hat{y}_{down})\|^2}$.
This allows us to visualize the rate of change in the high-dimensional space and quantitatively assess the inverse projection's gradients.
We compare the results across four datasets (\autoref{tab:datasets}) using t-SNE with standard parameters \cite{Maaten2008Tsne} as the ground-truth projection to be learned.

\begin{table}[t]
\setlength{\tabcolsep}{1.5mm}
\small
\centering
\begin{tabular}{lccccc}
\textbf{Dataset} & $\mathbf{d}$ & $\mathbf{\rho_d}$ & $\mathbf{n}$ & $\gamma_\mathbf{n}$ & \textbf{Type} \\
\hline
\textit{Rings} \cite{Blumberg2024} & $3$ & $3$ & $180$ & 0.0\% & Synthetic \\
\textit{HAR} \cite{Anguita2012} & $561$ & $120$ & $10299$ & 0.0\% & Sensor Data \\
\textit{MNIST} \cite{Lecun2010} & $784$ & $330$ & $70000$ & 82.9\% & Images \\
\textit{Fashion MNIST} \cite{Xiao2017} & $784$ & $187$ & $70000$ & 50.2\% & Images \\
\end{tabular}
\caption{Evaluation datasets with dimensionality $\mathbf{d}$, intrinsic dimensionality $\rho_\mathbf{d}$, dataset size $\mathbf{n}$, and sparsity $\gamma_\mathbf{n}$ \cite{Espadoto2019}.}
\label{tab:datasets}
\vspace*{-1em}
\end{table}

\begin{table}[t]
\setlength{\tabcolsep}{2.0mm}
\small
\centering
\begin{tabular}{ccccc}
& \textbf{Rings} & \textbf{HAR} & \textbf{MNIST} & \textbf{Fashion MNIST} \\
\hline
\hline
& \multicolumn{4}{c}{\textit{Average MSE of the Parametric Projection (lower is better)}} \\
\hline
\textbf{P\&R} & .119 (.021) & \textbf{.037 (.003)} & \textbf{.039 (.001)} & \textbf{.027 (.001)} \\
\textbf{AEL} & \textbf{.013 (.005)} & .041 (.002) & .074 (.004) & .046 (.003) \\
\textbf{VAEL} & .030 (.010) & .047 (.004) & .053 (.005) & .045 (.003) \\
\hline
& \multicolumn{4}{c}{\textit{Average MSE of the Inverse Projection (lower is better)}} \\
\hline
\textbf{P\&R} & .228 (.028) & .398 (.005) & \textbf{.720 (.010)} & \textbf{.459 (.008)} \\
\textbf{AEL} & \textbf{.033 (.012)} & \textbf{.357 (.006)} & .744 (.052) & .525 (.082) \\
\textbf{VAEL} & .118 (.037) & .420 (.005) & .847 (.048) & .556 (.019) \\
\end{tabular}
\caption{Aggregated MSE and standard deviation (in braces) of the parametric and inverse projections on test data for 10 runs each.
}
\label{tab:experiment-data}
\vspace*{-2em}
\end{table}

\begin{figure*}[t]
\newlength{\imgheight}
\setlength{\imgheight}{0.2\textwidth}
\setlength{\tabcolsep}{1.2mm}
\centering
\begin{tabular}{rllll}

& \horizontalLabel{Rings} & \horizontalLabel{HAR} & \horizontalLabel{MNIST} & \horizontalLabel{Fashion MNIST} \\

\verticalLabel{P\&R}&\includegraphics[height=\imgheight]{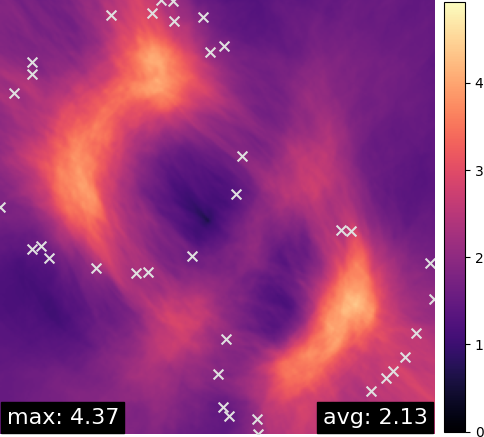} & \includegraphics[height=\imgheight]{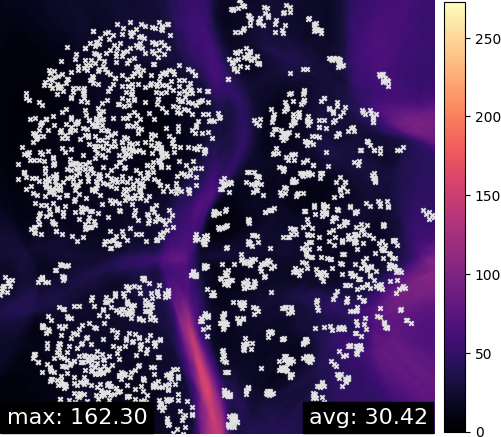} & \includegraphics[height=\imgheight]{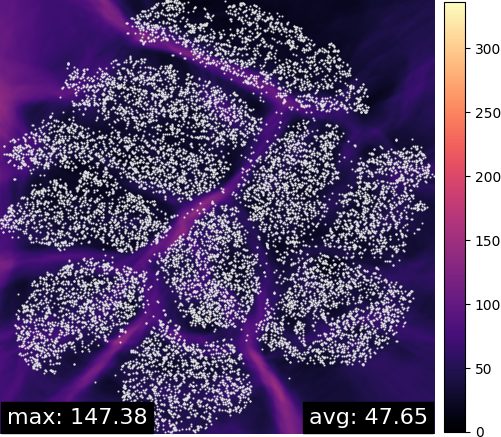} &\includegraphics[height=\imgheight]{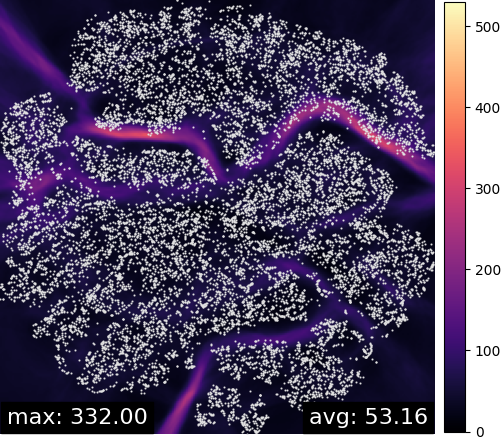} \\

\verticalLabel{AEL}&\includegraphics[height=\imgheight]{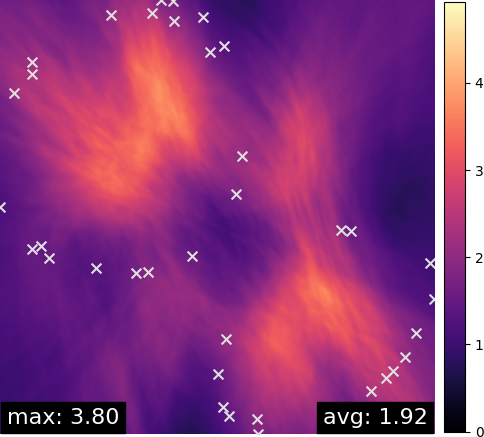} & \includegraphics[height=\imgheight]{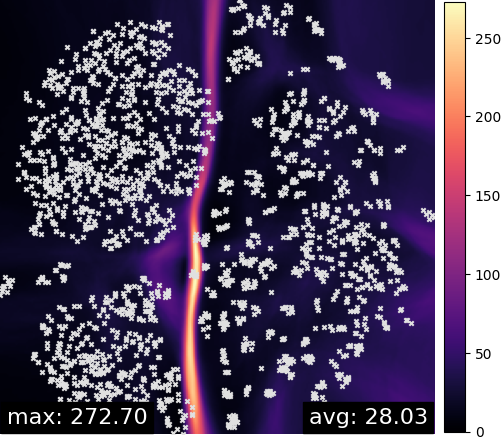} & \includegraphics[height=\imgheight]{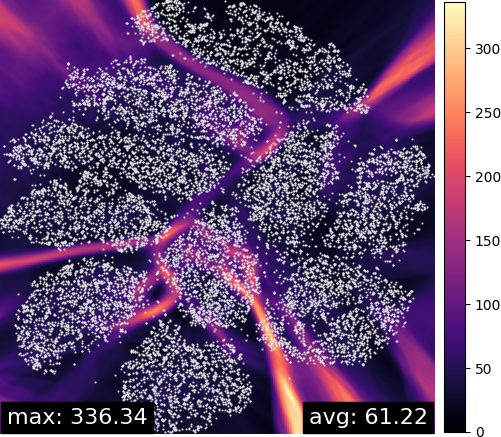} &\includegraphics[height=\imgheight]{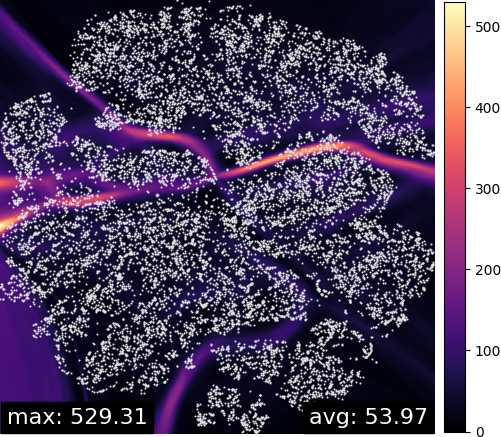} \\

\verticalLabel{VAEL}&\includegraphics[height=\imgheight]{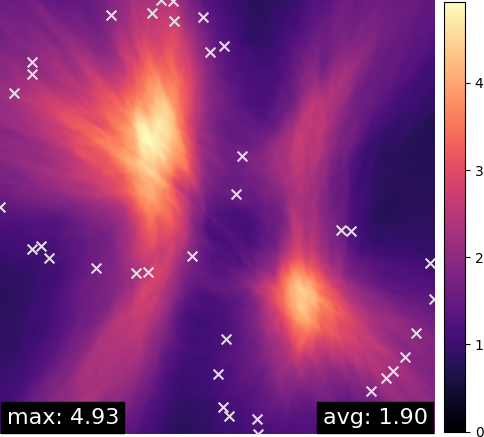} & \includegraphics[height=\imgheight]{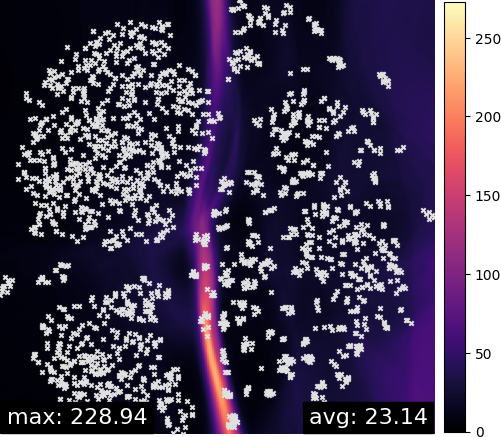} & \includegraphics[height=\imgheight]{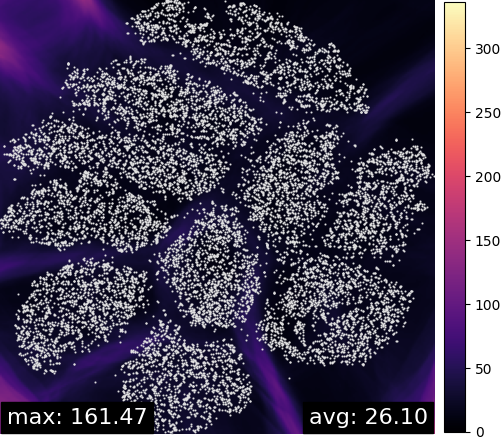} &\includegraphics[height=\imgheight]{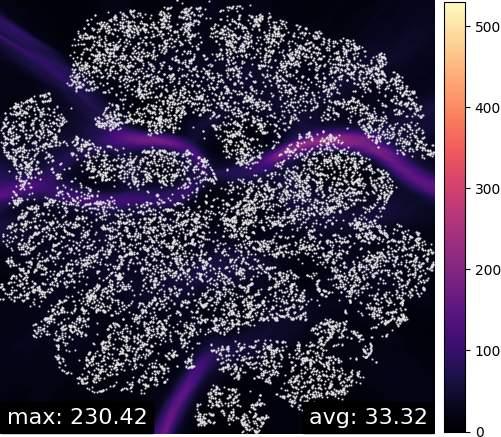} \\

\end{tabular}
\caption{Gradient maps of the three inverse projection methods for four datasets. Darker colors indicate a low rate of change, and lighter areas indicate a high rate of change. The numbers show the maximum gradient (bottom left) and average gradient (bottom right).}
\label{fig:gradient-maps}
\vspace*{-2em}
\end{figure*}

\subsection{Quantitative Comparison}

We evaluate the quality of the parametric and inverse projections by measuring the average MSE of test samples for each inverse method using an 80/20 train-test split.
To minimize the effect of outliers, we train 10 NNs for each method using different random assignments of items to the training and test sets.
The aggregated MSEs and their standard deviations are shown in \autoref{tab:experiment-data}.

\begin{figure}[tbhp]
\begin{minipage}[t]{0.23\linewidth}
\centering
\includegraphics[width=\linewidth]{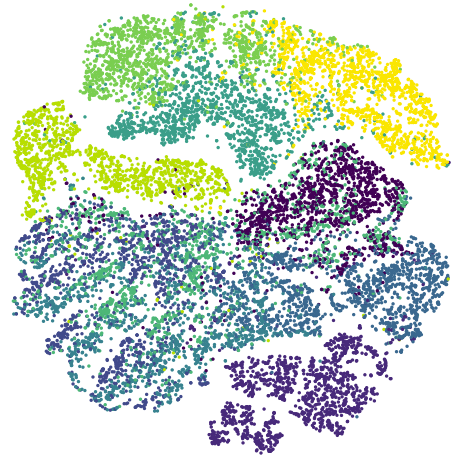} \\
(a) t-SNE
\end{minipage}
\hfill
\begin{minipage}[t]{0.23\linewidth}
\centering
\includegraphics[width=\linewidth]{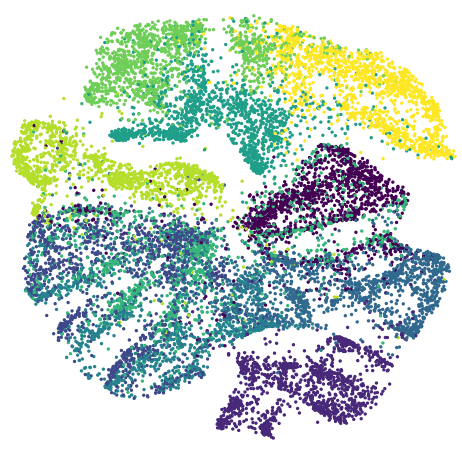} \\
(b) P\&R
\end{minipage}
\hfill
\begin{minipage}[t]{0.23\linewidth}
\centering
\includegraphics[width=\linewidth]{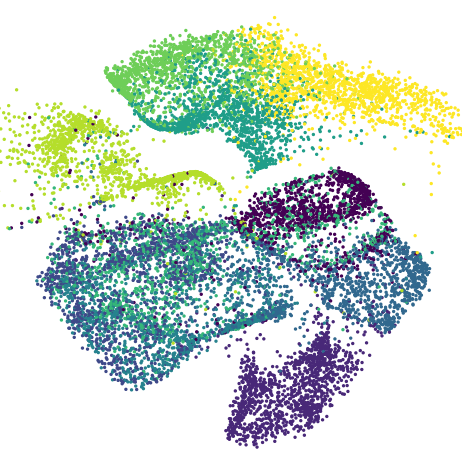} \\
(c) AEL
\end{minipage}
\hfill
\begin{minipage}[t]{0.23\linewidth}
\centering
\includegraphics[width=\linewidth]{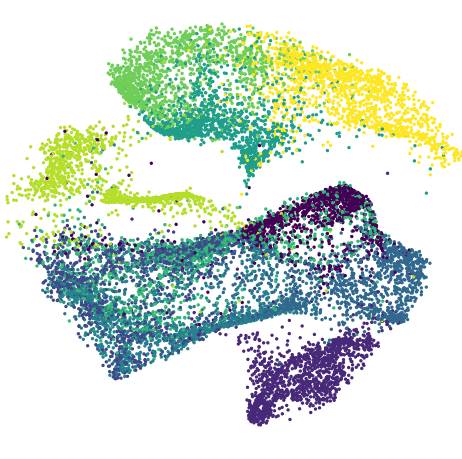} \\
(d) VAEL
\end{minipage}
\caption{Ground truth t-SNE projection of Fashion MNIST (a) and the parametric projections of test data for the three methods (b--d).}
\label{fig:parametric-projection}
\vspace*{-3em}
\end{figure}

\begin{figure*}[t]
    \centering
    \begin{minipage}[c]{0.71\linewidth}
        \vspace*{0mm}
        \setlength{\tabcolsep}{0.7mm}
        \begin{tabular}{rc}
            \verticalLabelSmall{\textbf{P\&R}} & \includegraphics[width=\linewidth]{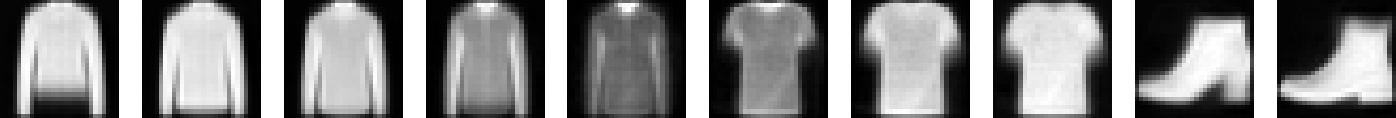} \\[0.3em]
            \verticalLabelSmall{\textbf{AEL}} & \includegraphics[width=\linewidth]{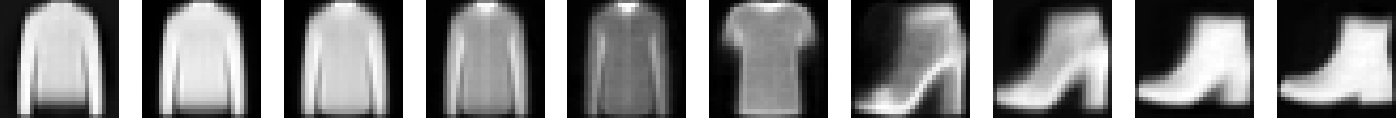} \\[0.3em]
            \verticalLabelSmall{\textbf{VAEL}} & \includegraphics[width=\linewidth,]{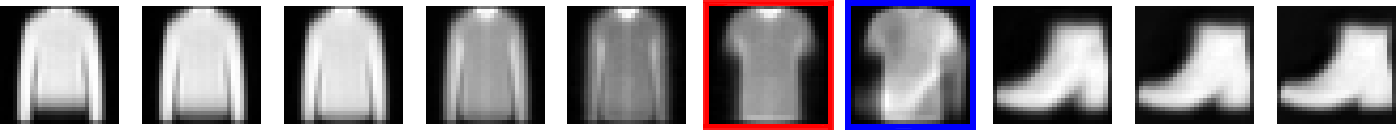}
        \end{tabular}
    \end{minipage}
    \hspace*{7mm}
    \begin{minipage}[c]{0.22\linewidth}
        \vspace*{0mm}
        \includegraphics[width=\linewidth,trim=7 7 7 7,clip]{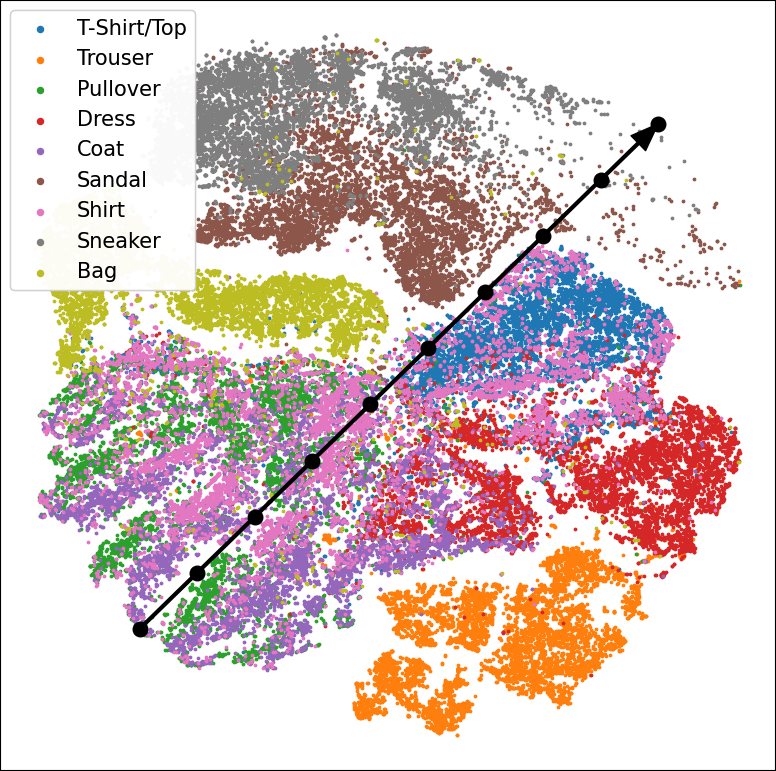}
    \end{minipage}
    \caption{For each of the three NN architectures, we inverse project 10 samples from an interpolation line of the projection space. Samples are drawn at equal intervals from an interpolation line in the t-SNE projection of the Fashion MNIST dataset (right).}
    \label{fig:interpolation}
    \vspace*{-2em}
\end{figure*}

The \textit{average MSE of the parametric projection} (\autoref{tab:experiment-data}, top) shows how well the different methods project high-dimensional data points for a given projection method $P$, i.e., t-SNE.
The P\&R applied to the \textit{Rings} dataset results in a high average MSE, suggesting that the projection performs the worst comparatively to the others, while AEL outperformed the other two.
For high-dimensional datasets (i.e., \textit{HAR}, \textit{MNIST}, and \textit{Fashion MNIST}), AEL and VAEL cannot outperform P\&R.
VAEL outperformed AEL in the \textit{MNIST} and \textit{Fashion MNIST} benchmarks.
The \textit{average MSE of the inverse projection} (\autoref{tab:experiment-data}, bottom) shows how well the approaches can create a high-dimensional data point from a 2D point in the projection.
AEL outperformed the other approaches for the \textit{Rings} and \textit{HAR} datasets.
For \textit{Fashion MNIST}, VAEL performed the worst.
Comparing P\&R with VAEL shows that P\&R generally outperformed VAEL except for the low-dimensional \textit{Rings} dataset.   
The \textit{average gradient} (\autoref{fig:gradient-maps}, bottom right)
measures the average rate of change of the high-dimensional inverse point when moving to a neighboring pixel in the 2D projection.
The average gradient increases with the intrinsic dimensionality of the dataset and generally decreases from P\&R to VAEL, suggesting that the VAEL generally produces smoother gradient maps.
\textit{Training times} generally increase with the intrinsic dimensionality and size of the dataset (see supplementary material).
P\&R requires the longest time to train, mainly because it involves training two separate networks. Additionally, the VAEL takes longer to train than the AEL.  
Overall, \textit{inference times} are similar and generally low for all the datasets, staying between 0.01s and 0.04s for all architectures.

\vspace{-1em}
\subsection{Qualitative Comparison}

We projected all samples in the test set using the learned parametric projections for the \textit{Fashion MNIST} dataset (\autoref{fig:parametric-projection}).
All parametric projections exhibit the general patterns of the t-SNE projection.
P\&R (b) looks similar to the ground truth.
The AEL projection (c) looks more clumpy and stringy than the P\&R projection.
VAEL (d) results in slightly fuzzier classes but misses the stringy artifacts in (c).
The observations confirm the quantitative evaluation.
We compared the inverse projection result for data records in the test set of the image datasets, i.e., \textit{MNIST} and \textit{Fashion MNIST}.
The results can be observed in the supplementary material.
P\&R and AEL produce output that can be assigned to one of the classes in the \textit{MNIST} and \textit{Fashion MNIST} datasets.
P\&R is more accurate than AEL since it produces more images matching the ground truth. 
VAEL will produce fuzzier-looking images compared to P\&R and AEL.
To show how the NNs handle transitions between multiple classes, we also visualized the inverse projections of samples of low-dimensional points along an interpolation line (\autoref{fig:interpolation}).
The samples generated by P\&R and AEL exhibit a generally high fidelity and can be assigned to a class of the \textit{Fashion MNIST} dataset.
The sample from VAEL marked red is less sharp, and the sample marked blue shows a shirt and a shadow of a boot.
Thus, VAEL produces smoother gradients while generating less accurate inverse projections.
Finally, we analyze the \textit{gradient maps} for the different datasets and NN architectures (\autoref{fig:gradient-maps}).
All approaches create differing gradient maps for the \textit{Rings} dataset, with AEL creating the smoothest map.
For the other datasets, the gradient maps show similar patterns, with higher gradients between clusters and generally lower gradients within clusters.
\textit{HAR} shows higher gradients on the right edge, while \textit{MNIST} exhibits a similar pattern on the left.
For the \textit{HAR}, a high gradient separates the two left clusters from the right in all maps.
For \textit{MNIST} and \textit{Fashion MNIST}, the maps show different gradients separating clusters.
P\&R creates smoother gradient maps for the higher-dimensional datasets than AEL with lower maximum gradients. 
When comparing the gradient maps of VAEL for \textit{HAR}, \textit{MNIST}, and \textit{Fashion MNIST} with the others, we can observe that the approach creates gradients in similar locations.
However, their average gradients are the lowest.
Thus, the inverse projections of VAEL are generally \emph{smoother}.

\section{Discussion and Future Work}\label{sec:discussion}

The results for P\&R show that a feed-forward NN generally works best for creating a parametric projection.
Similarly, P\&R outperformed AEL and VAEL for generating inverse projections of \textit{MNIST} and \textit{Fashion MNIST}.
The architectures of AEL and VAEL are directly derived from P\&R.
However, AEL and VAEL may benefit from additional and larger layers, requiring a validation set to determine their hyperparameters.
Our results suggest that the \textit{joint} training of $P$ and $P^{-1}$ using AEs generally yields \textit{smoother} inverse projections.
In terms of smoothness, VAEL gives the best results with low average gradients and smaller areas of high gradients (\autoref{fig:gradient-maps}), showing that incorporating the $D_{KL}$ loss term has a smoothing effect.
Our evaluation suggests that this effect comes at the cost of the parametric and inverse projection accuracy.
However, the smoothing strength is controllable through the parameters of VAEL.

\subhead{Loss Weights $\omega$, $\alpha$, and $\beta$:} We determined the values for $\omega$, $\alpha$, and $\beta$ experimentally by calculating the MSE on reconstruction and latent space for different levels of $\omega$, $\alpha$, and $\beta$ (see supplementary material). For AEL, choosing a lower $\omega$ leads to a lower MSE on inverse projection and a higher MSE on the parametric projection. We recommend selecting an $\omega$ of $0.5$ as the values between $0.4$ and $5.0$ yield similarly low MSEs. 
For the VAEL, the smaller the $\beta$, the better the weighted sum of MSE on parametric and inverse projection. In contrast, the weighted sum decreases with larger $\alpha$. We chose a $\beta$ of $0.1$ and an $\alpha$ of $1.0$ since these weights yield generally low MSE losses. $\alpha$ can be increased to $5.0$ (with $\beta = 0.1$), without negatively affecting the MSE. $\beta$ is the more sensitive parameter directly influencing the smoothness of the inverse projection.
We recommend evaluating the $\beta$ parameter on a case-by-case basis.

\subhead{Future Work:}
Our work focused on t-SNE; however, we plan to extend the evaluation to other projection methods, like UMAP.
We also plan to compare our AE-based architectures to existing parametric and inverse projection methods.
Since we used common datasets and an 80/20 train-test split, we would like to further test the stability of each approach w.r.t.~amount of training data and intrinsic dimensionality.  
In our approach, $D_{KL}$ of the VAE serves as an implicit regularization of the latent space.
However, a similar effect could be achieved with a standard AE using explicit regularization, such as an L2-regularization of the Jacobian of the latent layer.
We adapted network topologies from previous applications, but we would like to perform ablation studies to assess the degradation of the proposed architectures.
Finally, other NN types could be tested, namely generative adversarial networks (GANs) or invertible NNs.

\section{Conclusion}

We evaluated three different AEs for generating \textit{parametric} and \textit{invertible} multidimensional data projections. 
Our qualitative and quantitative results for t-SNE showed the differences in projection and reconstruction capabilities between the tested models. 
In general, we found that feed-forward NNs for the projection and reconstruction of data points generally outperform AE-based approaches in terms of accuracy. 
However, AEs can produce comparable results while generally producing smoother parametric and inverse projections.
In particular, we found that VAEs with a customized loss function have the greatest potential for producing smooth inverse projections.
Their parameterization allows for case-by-case fine-tuning between overall accuracy and smoothness.

\noindent
\textbf{Acknowledgments --}
This work was funded by the Deutsche Forschungsgemeinschaft (DFG, German Research Foundation) -- Project-ID 251654672 -- TRR 161 (Project A03).



\printbibliography           

@inproceedings{Blumberg2024,
  booktitle  = {15th Int. EuroVis Workshop Vis. Anal.},
  _editor    = {El-Assady, Mennatallah and Schulz, Hans-Jörg},
  title      = {{Inverting Multidimensional Scaling Projections Using Data Point Multilateration}},
  author     = {Blumberg, Daniela and Wang, Yu and Telea, Alexandru and Keim, Daniel A. and Dennig, Frederik L.},
  year       = {2024},
  _publisher = {The Eurographics Association},
  _isbn      = {978-3-03868-253-0},
  doi        = {10.2312/eurova.20241112}
}

@inproceedings{Wijk2003,
  author       = {Jarke J. van Wijk and
                  Cornelius W. A. M. van Overveld},
  _editor      = {Frits H. Post and
                  Gregory M. Nielson and
                  Georges{-}Pierre Bonneau},
  title        = {Preset based interaction with high dimensional parameter spaces},
  booktitle    = {Data Visualization: The State of the Art},
  pages        = {391--406},
  _publisher   = {Kluwer},
  year         = {2003}
}

@article{McInnes2018,
  author       = {Leland McInnes and
                  John Healy and
                 James Melville},
  title        = {{UMAP: Uniform Manifold Approximation and Projection for Dimension
                  Reduction}},
  journal      = {CoRR},
  volume       = {abs/1802.03426},
  year         = {2018},
  _url          = {http://arxiv.org/abs/1802.03426},
  _eprinttype    = {arXiv},
  _eprint       = {1802.03426},
  _doi          = {10.48550/arXiv.1802.03426}
}

@inproceedings{Amorim2012,
  author       = {Elisa Portes dos Santos Amorim and
                  Emilio Vital Brazil and
                  Joel Daniels II and
                  Paulo Joia and
                  Luis Gustavo Nonato and
                  Mario Costa Sousa},
  title        = {{iLAMP}: Exploring high-dimensional spaces through backward multidimensional projection},
  booktitle    = {7th {IEEE} Conf. Vis. Anal. Sci. Technol.},
  pages        = {53--62},
  _publisher   = {{IEEE} Computer Society},
  year         = {2012},
  _url         = {https://doi.org/10.1109/VAST.2012.6400489},
  doi          = {10.1109/VAST.2012.6400489}
}

@article{Amorim2015,
  author       = {Elisa Amorim and
                  Emilio Vital Brazil and
                  Jes{\'{u}}s Mena{-}Chalco and
                  Luiz Velho and
                  Luis Gustavo Nonato and
                  Faramarz Samavati and
                  Mario Costa Sousa},
  title        = {Facing the high-dimensions: Inverse projection with radial basis functions},
  journal      = {Comput. Graph.},
  volume       = {48},
  pages        = {35--47},
  year         = {2015},
  url          = {https://doi.org/10.1016/j.cag.2015.02.009},
  doi          = {10.1016/J.CAG.2015.02.009}
}

@article{Espadoto2019,
  author       = {Mateus Espadoto and
                  Rafael Messias Martins and
                  Andreas Kerren and
                  Nina S. T. Hirata and
                  Alexandru C. Telea},
  title        = {Toward a Quantitative Survey of Dimension Reduction Techniques},
  journal      = {{IEEE} Trans. Vis. Comput. Graph.},
  volume       = {27},
  number       = {3},
  pages        = {2153--2173},
  year         = {2019},
  _url         = {https://doi.org/10.1109/TVCG.2019.2944182},
  doi          = {10.1109/TVCG.2019.2944182}
}

@article{Nonato2019,
  author       = {Luis Gustavo Nonato and
                  Micha{\"{e}}l Aupetit},
  title        = {Multidimensional Projection for Visual Analytics: Linking Techniques
                  with Distortions, Tasks, and Layout Enrichment},
  journal      = {{IEEE} Trans. Vis. Comput. Graph.},
  volume       = {25},
  number       = {8},
  _pages       = {2650--2673},
  year         = {2019},
  _url         = {https://doi.org/10.1109/TVCG.2018.2846735},
  doi          = {10.1109/TVCG.2018.2846735}
}

@article{Cunningham2015,
  author       = {John P. Cunningham and
                  Zoubin Ghahramani},
  title        = {Linear dimensionality reduction: survey, insights, and generalizations},
  journal      = {J. Mach. Learn. Res.},
  volume       = {16},
  pages        = {2859--2900},
  year         = {2015},
  _url         = {https://dl.acm.org/doi/10.5555/2789272.2912091},
  doi          = {10.5555/2789272.2912091}
}

@article{Shusen2017,  
  author       = {Shusen Liu and
                  Dan Maljovec and
                  Bei Wang and
                  Peer{-}Timo Bremer and
                  Valerio Pascucci},
  title        = {Visualizing High-Dimensional Data: Advances in the Past Decade},
  journal      = {{IEEE} Trans. Vis. Comput. Graph.},
  volume       = {23},
  number       = {3},
  pages        = {1249--1268},
  year         = {2017},
  _url         = {https://doi.org/10.1109/TVCG.2016.2640960},
  doi          = {10.1109/TVCG.2016.2640960}
}

@article{kruskal1978multidimensional,
  title={Multidimensional scaling},
  author={Kruskal, Joseph B},
  journal={Murry Hill},
  year={1978}
}

@article{Yin2007,
  author       = {Hujun Yin},
  title        = {Nonlinear dimensionality reduction and data visualization: {A} review},
  journal      = {Int. J. Autom. Comput.},
  volume       = {4},
  number       = {3},
  pages        = {294--303},
  year         = {2007},
  _url         = {https://doi.org/10.1007/s11633-007-0294-y},
  doi          = {10.1007/S11633-007-0294-Y},
}

@book{Jolliffe1986,
  author    = {Ian T. Jolliffe},
  title     = {Principal Component Analysis},
  _series   = {Springer Series in Statistics},
  publisher = {Springer},
  year      = {1986},
  _url      = {https://doi.org/10.1007/978-1-4757-1904-8},
  doi       = {10.1007/978-1-4757-1904-8},
  _isbn     = {978-1-4757-1906-2}
}

@article{Hinterreiter2023,
  author       = {Andreas P. Hinterreiter and
                  Christina Humer and
                  Bernhard Kainz and
                  Marc Streit},
  title        = {ParaDime: {A} Framework for Parametric Dimensionality Reduction},
  journal      = {Comput. Graph. Forum},
  volume       = {42},
  number       = {3},
  pages        = {337--348},
  year         = {2023},
  _url         = {https://doi.org/10.1111/cgf.14834},
  doi          = {10.1111/CGF.14834}
}

@article{Sainburg2021,
  author       = {Tim Sainburg and
                  Leland McInnes and
                  Timothy Q. Gentner},
  title        = {Parametric {UMAP} Embeddings for Representation and Semisupervised
                  Learning},
  journal      = {Neural Comput.},
  volume       = {33},
  number       = {11},
  pages        = {2881--2907},
  year         = {2021},
  _url         = {https://doi.org/10.1162/neco\_a\_01434},
  doi          = {10.1162/NECO\_A\_01434}
}

@article{wang2016auto,
  title={Auto-encoder based dimensionality reduction},
  author={Wang, Yasi and Yao, Hongxun and Zhao, Sicheng},
  journal={Neurocomputing},
  volume={184},
  pages={232--242},
  year={2016},
  _publisher={Elsevier}
}

@article{Joia2011,
  author       = {Paulo Joia and
                  Danilo Barbosa Coimbra and
                  Jos{\'{e}} Alberto Cuminato and
                  Fernando Vieira Paulovich and
                  Luis Gustavo Nonato},
  title        = {Local Affine Multidimensional Projection},
  journal      = {{IEEE} Trans. Vis. Comput. Graph.},
  volume       = {17},
  number       = {12},
  pages        = {2563--2571},
  year         = {2011},
  _url         = {https://doi.org/10.1109/TVCG.2011.220},
  doi          = {10.1109/TVCG.2011.220}
}

@inproceedings{Higgins2017,
  author       = {Irina Higgins and
                  Loic Matthey and
                  Arka Pal and
                  Christopher P. Burgess and
                  Xavier Glorot and
                  Matthew M. Botvinick and
                  Shakir Mohamed and
                  Alexander Lerchner},
  title        = {beta-VAE: Learning Basic Visual Concepts with a Constrained Variational
                  Framework},
  booktitle    = {5th Int. Conf. Learn. Represent.},
  _publisher    = {OpenReview.net},
  year         = {2017},
  _url         = {https://openreview.net/forum?id=Sy2fzU9gl}
}

@article{Espadoto2020Deep,
  author       = {Mateus Espadoto and
                  Nina Sumiko Tomita Hirata and
                  Alexandru C. Telea},
  title        = {Deep learning multidimensional projections},
  journal      = {Inf. Vis.},
  volume       = {19},
  number       = {3},
  pages        = {247--269},
  year         = {2020},
  _url         = {https://doi.org/10.1177/1473871620909485},
  doi          = {10.1177/1473871620909485}
}

@article{Appleby2022,
  author       = {Gabriel Appleby and
                  Mateus Espadoto and
                  Rui Chen and
                  Samuel Goree and
                  Alexandru C. Telea and
                  Erik W. Anderson and
                  Remco Chang},
  title        = {{HyperNP}: Interactive Visual Exploration of Multidimensional Projection
                  Hyperparameters},
  journal      = {Comput. Graph. Forum},
  volume       = {41},
  number       = {3},
  pages        = {169--181},
  year         = {2022},
  _url         = {https://doi.org/10.1111/cgf.14531},
  doi          = {10.1111/CGF.14531}
}

@article{Espadoto2021Unprojection,
  author       = {Mateus Espadoto and
                  Gabriel Appleby and
                  Ashley Suh and
                  Dylan Cashman and
                  Mingwei Li and
                  Carlos Scheidegger and
                  Erik W. Anderson and
                  Remco Chang and
                  Alexandru C. Telea},
  title        = {{UnProjection}: Leveraging Inverse-Projections for Visual Analytics
                  of High-Dimensional Data},
  journal      = {IEEE Trans. Vis. Comput. Graph.},
  volume       = {29},
  number       = {2},
  pages        = {1559--1572},
  year         = {2021},
  _url         = {https://doi.org/10.1109/TVCG.2021.3125576},
  doi          = {10.1109/TVCG.2021.3125576}
}

@article{Hinton2006,
  author  = {G. E. Hinton  and R. R. Salakhutdinov },
  title   = {Reducing the Dimensionality of Data with Neural Networks},
  journal = {Science},
  volume  = {313},
  number  = {5786},
  pages   = {504--507},
  year    = {2006},
  doi     = {10.1126/science.1127647},
  _url    = {https://www.science.org/doi/abs/10.1126/science.1127647},
  _eprint = {https://www.science.org/doi/pdf/10.1126/science.1127647}
}

@article{Machado2024,
  author       = {Alister Machado and
                  Alexandru C. Telea and
                  Michael Behrisch},
  title        = {Controlling the scatterplot shapes of 2D and 3D multidimensional projections},
  journal      = {Comput. Graph.},
  volume       = {124},
  pages        = {104093},
  year         = {2024},
  _url         = {https://doi.org/10.1016/j.cag.2024.104093},
  doi          = {10.1016/J.CAG.2024.104093}
}

@inproceedings{Maaten2009,
  author       = {Laurens van der Maaten},
  _editor      = {David A. Van Dyk and
                  Max Welling},
  title        = {Learning a Parametric Embedding by Preserving Local Structure},
  booktitle    = {12th Int. Conf. Artif. Intell. Stat.},
  _series      = {{JMLR} Proceedings},
  volume       = {5},
  pages        = {384--391},
  _publisher   = {JMLR.org},
  year         = {2009},
  _url         = {http://proceedings.mlr.press/v5/maaten09a.html},
}

@article{Maaten2008Tsne,
  author  = {Laurens van der Maaten and Geoffrey Hinton},
  title   = {Visualizing Data using t-SNE},
  journal = {J. Mach. Learn. Res.},
  year    = {2008},
  volume  = {9},
  number  = {86},
  pages   = {2579--2605},
  _url    = {http://jmlr.org/papers/v9/vandermaaten08a.html}
}

@inproceedings{Espadoto2021Ssnp,
  author       = {Mateus Espadoto and
                  Nina S. T. Hirata and
                  Alexandru C. Telea},
  _editor      = {Christophe Hurter and
                  Helen C. Purchase and
                  Jos{\'{e}} Braz and
                  Kadi Bouatouch},
  title        = {Self-supervised Dimensionality Reduction with Neural Networks and
                  Pseudo-labeling},
  booktitle    = {16th Int. Jt. Conf. Comput. Vis. Imaging Comput. Graph. Theory Appl.},
  pages        = {27--37},
  _publisher   = {{SCITEPRESS}},
  year         = {2021},
  _url         = {https://doi.org/10.5220/0010184800270037},
  doi          = {10.5220/0010184800270037}
}

@article{Dennig2024,
  author       = {Frederik L. Dennig and
                  Matthias Miller and
                  Daniel A. Keim and
                  Mennatallah El{-}Assady},
  title        = {{FS/DS:} {A} Theoretical Framework for the Dual Analysis of Feature
                  Space and Data Space},
  journal      = {{IEEE} Trans. Vis. Comput. Graph.},
  volume       = {30},
  number       = {8},
  pages        = {5165--5182},
  year         = {2024},
  _url         = {https://doi.org/10.1109/TVCG.2023.3288356},
  doi          = {10.1109/TVCG.2023.3288356}
}

@article{Xiao2017,
  author       = {Han Xiao and
                  Kashif Rasul and
                  Roland Vollgraf},
  title        = {Fashion-MNIST: a Novel Image Dataset for Benchmarking Machine Learning
                  Algorithms},
  journal      = {CoRR},
  volume       = {abs/1708.07747},
  year         = {2017},
  _url         = {http://arxiv.org/abs/1708.07747},
  _eprinttype   = {arXiv},
  _eprint       = {1708.07747}
}

@inproceedings{Kingma2015,
  author       = {Diederik P. Kingma and
                  Jimmy Ba},
  _editor      = {Yoshua Bengio and
                  Yann LeCun},
  title        = {Adam: {A} Method for Stochastic Optimization},
  booktitle    = {3rd Int. Conf. Learn. Represent.},
  year         = {2015},
  _url          = {http://arxiv.org/abs/1412.6980}
}

@inproceedings{Kingma2014,
  author      = {Kingma, Diederik P. and Welling, Max},
  booktitle   = {2nd Int. Conf. Learn. Represent.},
  _eprint      = {http://arxiv.org/abs/1312.6114v10},
  _eprintclass = {stat.ML},
  _eprinttype  = {arXiv},
  title       = {{Auto-Encoding Variational Bayes}},
  year        = {2014}
}

@inproceedings{Anguita2012,
  title        = {Human activity recognition on smartphones using a multiclass hardware-friendly support vector machine},
  author       = {Anguita, Davide and Ghio, Alessandro and Oneto, Luca and Parra, Xavier and Reyes-Ortiz, Jorge L},
  booktitle    = {Ambient Assist. Living Home Care},
  pages        = {216--223},
  year         = {2012},
  organization = {Springer}
}

@misc{Lecun2010,
  title     = {{MNIST} handwritten digit database},
  author    = {LeCun, Yann and Cortes, Corinna and Burges, Chris},
  year      = {1998},
}

@Inbook{Bank2023,
  author    = {Bank, Dor and Koenigstein, Noam and Giryes, Raja},
  _editor   = {Rokach, Lior and Maimon, Oded and Shmueli, Erez},
  title     = {Autoencoders},
  booktitle = {Machine Learning for Data Science Handbook: Data Mining and Knowledge Discovery Handbook},
  year      = {2023},
  publisher = {Springer},
  _address  = {Cham},
  pages     = {353--374},
  _isbn     = {978-3-031-24628-9},
  doi       = {10.1007/978-3-031-24628-9_16},
  _url      = {https://doi.org/10.1007/978-3-031-24628-9_16}
}

@inproceedings{Chen2024,
  author       = {Florian Chen and
                  Thomas G{\"{a}}rtner},
  _editor      = {Albert Bifet and
                  Povilas Daniusis and
                  Jesse Davis and
                  Tomas Krilavicius and
                  Meelis Kull and
                  Eirini Ntoutsi and
                  Kai Puolam{\"{a}}ki and
                  Indre Zliobaite},
  title        = {Scalable Interactive Data Visualization},
  booktitle    = {Mach. Learn. Knowl. Discov. Databases},
  _series      = {Lecture Notes in Computer Science},
  volume       = {14948},
  pages        = {429--433},
  _publisher   = {Springer},
  year         = {2024},
  _url          = {https://doi.org/10.1007/978-3-031-70371-3\_34},
  doi          = {10.1007/978-3-031-70371-3\_34}
}

@article{Bunte2012,
  author       = {Kerstin Bunte and
                  Michael Biehl and
                  Barbara Hammer},
  title        = {A General Framework for Dimensionality-Reducing Data Visualization
                  Mapping},
  journal      = {Neural Comput.},
  volume       = {24},
  number       = {3},
  pages        = {771--804},
  year         = {2012},
  _url         = {https://doi.org/10.1162/NECO\_a\_00250},
  doi          = {10.1162/NECO\_A\_00250}
}

@inproceedings{Schlegel2024,
  author       = {Udo Schlegel and
                  Julius Rauscher and
                  Daniel A. Keim},
  booktitle    = {6th Int. Workshop Explain. Knowl. Discov. Data Min.},
  title        = {Interactive Counterfactual Generation for Univariate Time Series},
  year         = {2024}
}


\end{document}